\documentclass[runningheads]{llncs}
\usepackage{graphicx}
\usepackage{comment}
\usepackage{amsmath,amssymb} 
\usepackage{color}
\usepackage{amsmath}
\usepackage{amsmath,amssymb}
\usepackage{multirow}
\usepackage{array}
\usepackage{hyperref}

\emergencystretch=\maxdimen
\begin{document}
\pagestyle{headings}
\mainmatter
\def\ECCVSubNumber{2865}  

\title{Open-set Adversarial Defense} 



\titlerunning{Open-set Adversarial Defense}

\author{Rui Shao\inst{1}\orcidID{0000-0003-0090-9604} \and
	Pramuditha Perera\inst{2}\orcidID{0000-0003-2821-6367}\thanks{This work was conducted prior to joining AWS AI Labs when the  author was affiliated with Johns Hopkins University.} \and \\
	Pong C. Yuen\inst{1}\orcidID{0000-0002-9343-2202}\and
	Vishal M. Patel\inst{3}\orcidID{0000-0002-5239-692X}
}

\authorrunning{Rui Shao, Pramuditha Perera, Pong C. Yuen, Vishal M. Patel}

\institute{$^1$Department of Computer Science, Hong Kong Baptist University, Hong Kong\\
		$^2$AWS AI Labs, USA\\
	$^3$Department of Electrical and Computer Engineering, Johns Hopkins University, USA\\
	\email{\{ruishao, pcyuen\}@comp.hkbu.edu.hk, pramudi@amazon.com, vpatel36@jhu.edu}}
\maketitle

\begin{abstract}
Open-set recognition and adversarial defense study two key aspects of deep learning that are vital for real-world deployment. The objective of open-set recognition is to identify samples from open-set classes during testing, while adversarial defense aims to defend the network against images with imperceptible adversarial perturbations. In this paper, we show that open-set recognition systems are vulnerable to adversarial attacks.  Furthermore, we show that adversarial defense mechanisms trained on known classes do not generalize well to open-set samples. Motivated by this observation, we emphasize the need of an Open-Set Adversarial Defense (OSAD) mechanism. This paper proposes an Open-Set Defense Network (OSDN) as a solution to the OSAD problem.  The proposed network uses an encoder with feature-denoising layers coupled with a classifier to learn a noise-free latent feature representation. Two techniques are employed to obtain an informative latent feature space with the objective of improving open-set performance. First, a decoder is used to ensure that clean images can be reconstructed from the obtained latent features. Then, self-supervision is used to ensure that the latent features are informative enough to carry out an auxiliary task. We introduce a testing protocol to evaluate OSAD performance and show the effectiveness of the proposed method in multiple object classification datasets. The implementation code of the proposed method is available at: \href{https://github.com/rshaojimmy/ECCV2020-OSAD}{https://github.com/rshaojimmy/ECCV2020-OSAD}.

\end{abstract}

\section{Introduction}

\begin{figure}[t] 
	
	\begin{center}
		
		\includegraphics[height=3cm, width=1\linewidth]{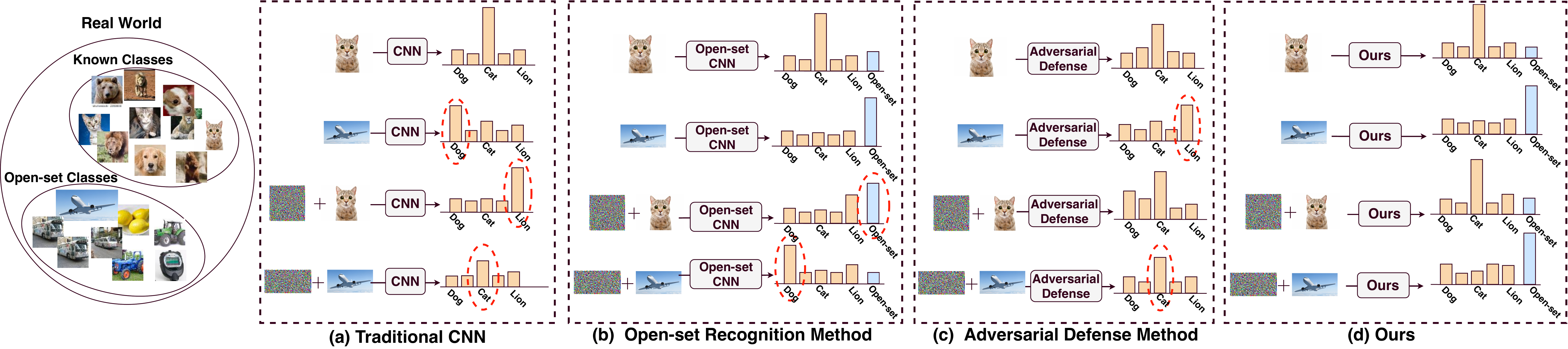}
		
	\end{center}
	
	\caption{Challenges in classification. (a) Conventional CNN classifiers fail in the presence of both open-set and adversarial images. (b) Open-set recognition methods can successfully identify open-set samples, but fail on adversarial samples. (c) Adversarial defense methods are unable to identify open-set samples. (d) Proposed method can identify open-set images and it is robust to adversarial images.}
	\label{fig:overview}
\end{figure}

\begin{table}[t] 
	\centering
	\label{tbl:init}
	\caption{Importance of an Open-set Adversarial Defense (OSAD) mechanism.}                
		\resizebox{0.8\linewidth}{!}{
	\begin{tabular}{|c|c|c|c|l|}
		\hline
		& Clean Images     & \multicolumn{3}{c|}{Adversarial Images}                \\ \hline
		& Original Network & Original Network       & \multicolumn{2}{c|}{Proposed Method} \\ \hline
		Closed Set Accuracy          & 96.0             & 31.8                   & \multicolumn{2}{c|}{88.2}     \\ \hline
		Open-set Detection (AUC-ROC) & 81.2             & 51.5                   & \multicolumn{2}{c|}{79.1}     \\ \hline
	\end{tabular}
}
\end{table}

A significant improvement has been achieved in the image classification task since the advent of deep convolutional neural networks (CNNs)~\cite{Kaiming_Resnet_CVPR2016}. The promising performance in classification has contributed to many real-world computer vision applications~\cite{shao2020regularized,Shao2019CVPR,2018TIFSdynamictext,RuiShao2018IJCB,2018MMhada,Shao_2019_IETIP,shao2020federated,arxiv20reidsurvey,cvpr19uel,Tracking_2019_TIE,Yash2019anomaly,pami20isif}. However, there exist several limitations of conventional CNNs that have an impact in real-world applications. In particular, open-set recognition~\cite{bendale2016towards,ge2017generative,neal2018open,oza2019c2ae,yoshihashi2019classification,Perera_2020_CVPR,Perera2019Noveltytransfer,Perera_CVPR19_2,zhang2016sparse} and adversarial attacks~\cite{goodfellow2014explaining,madry2017towards,carlini2017towards,kurakin2016adversarial,xie2019feature} have received a lot of interest in the computer vision community in the last few years.  

A classifier is conventionally trained assuming that classes encountered during testing will be identical to classes observed during training. But in a real-world scenario, a trained classifier is likely to encounter open-set samples from classes unseen during training. When this is the case, the classifier will erroneously associate a known-set class identity to an open-set sample. Consider a CNN trained on animals classes. Given an input that is from an animal class (such as a cat), the network is able to produce the correct prediction as shown in Figure~\ref{fig:overview}(a-First Row). However, when the network is presented with a non-animal image, such as an Airplane image, the classifier wrongly classifies it as one of the known classes as shown in Figure~\ref{fig:overview}(a-Second Row). On the other hand, it is a well known fact that adding carefully designed imperceptible perturbations to clean images can alter model prediction in a classifier~\cite{goodfellow2014explaining}. These types of \textit{adversarial attacks} are easy to deploy and may be encountered in real-world applications~\cite{evtimov2017robust}. In Figure~\ref{fig:overview}(a-Third Row) and Figure~\ref{fig:overview}(a-Fourth Row), we show how such adversarial attacks can affect model prediction for known and open-set images, respectively.

Computer vision community has developed several open-set recognition algorithms~\cite{bendale2016towards,ge2017generative,neal2018open,oza2019c2ae,yoshihashi2019classification} to combat against the former challenge. These algorithms convert the $c$-class classification problem into a $c+1$ class problem by considering open-set classes as an additional class. These algorithms provide correct classification decisions for both known and open-set classes as shown in Figure~\ref{fig:overview}(b-First and Second rows). However, in the presence of adversarial attacks, these models fail to produce correct predictions as illustrated in Figure~\ref{fig:overview}(b-Third and Fourth rows). On the other hand, there exist several defense strategies~\cite{kurakin2016adversarial,xie2019feature,liao2018defense,jang2019adversarial} that are proposed to counter the latter challenge. These defense mechanisms are designed with the assumption of closed-set testing. Therefore, although they work well when this assumption holds (Figure~\ref{fig:overview}(c-First and third rows)), they fail to generalize well in the presence of open-set samples as shown in Figure~\ref{fig:overview}(c-Second and Fourth rows).

Based on this discussion, it is evident that existing solutions in the open-set recognition paradigm does not necessarily complement well with adversarial defense and vise versa. This observation motivates us to introduce a new research problem -- Open-Set Adversarial Defense (OSAD), where the objective is to simultaneously detect open-set samples and classify known classes in the presence of adversarial noise. In order to demonstrate the significance of the proposed problem, we conducted an experiment on the CIFAR10 dataset by considering only 6 classes to be known to the classifier. In Table~\ref{tbl:init} we tabulate both open-set detection performance (expressed in terms of area under the curve of the ROC curve) and closed-set classification accuracy for this experiment. When the network is presented with clean images, it produces a performance better than 0.8 in both open-set detection and closed set classification. However, as evident from Table~\ref{tbl:init},  when images are attacked, open-set detection performance drops along with the closed set accuracy by a significant margin. It should be noted that, open-set detection performance in this case is close to random guessing $(0.5)$.

This paper proposes an Open-Set Defense Network (OSDN) that learns a noise-free, informative latent feature space with the aim of detecting open-set samples while being robust to adversarial images. We use an autoencoder network with a classifier branch attached to its latent space as the backbone of our solution. The encoder network is equipped with feature-denoising layers~\cite{xie2019feature} with the aim of removing adversarial noise. We use self-supervision and decoder reconstruction processes to make sure that the learned feature space is informative enough to detect open-set samples. The reconstruction process uses the decoder to generate noise-free images based on the obtained latent features. Self-supervision is carried out by forcing the network to perform an auxiliary classification task based on the obtained features. The proposed method is able to provide robustness against adversarial attacks in terms of classification as well as open-set detection as shown in Table~\ref{tbl:init}. Main contributions of our paper can be summarized as follows:

\noindent 1. This paper proposes a new research problem named Open-Set Adversarial Defense (OSAD) where adversarial attacks are studied under an open-set setting.

\noindent 2. We propose an Open-Set Defense Network (OSDN) that learns a latent feature space that is robust to adversarial attacks and informative to identify open-set samples.

\noindent3. A test protocol is defined to the OSAD problem. Extensive experiments are conducted on three publicly available image classification datasets to demonstrate the effectiveness of the proposed method.

\section{Related Work}

\noindent \textbf{Adversarial Attack and Defense Methods.}
Szegedy \textit{et al.}~\cite{szegedy2013intriguing} reported that carefully crafted  imperceptible perturbations can be used to fool a CNN to make incorrect predictions. Since then, various adversarial attacks have been proposed in the literature. Fast Gradient Sign Method (FGSM)~\cite{goodfellow2014explaining} was proposed to consider the sign of a gradient update from the classifier to generate adversarial images. Basic Iteration Method (BIM)~\cite{kurakin2016adversarial} and Projected Gradient Descent (PGD)~\cite{madry2017towards} extended FGSM to stronger attacks using iterative gradient descent. Different from the above gradient-based adversarial attacks, Carlini and Wagner~\cite{carlini2017towards} proposed the C$\&$W attack to generate adversarial samples by taking a direct optimization approach. Adversarial training~\cite{madry2017towards} is one of the most widely-used adversarial defense mechanisms. It provides defense against adversarial attacks by training the network on adversarially perturbed images generated on-the-fly based on model's current parameters. Several recent works have proposed denoising-based operations to further improve adversarial training. Pixel denoising~\cite{liao2018defense} was proposed to exploit the high-level features to guide the denoising process. The most influential local parts to conduct the pixel-level denoising is found in~\cite{gupta2019ciidefence} based on class activation map responses. Adversarial noise removal is carried out in the feature-level using denoising filters in~\cite{xie2019feature}. Effectiveness of this process is demonstrated using a selection of different filters.

\noindent \textbf{Open-set Recognition.} Possibility for open-set samples to generate very high probability scores in a closed-set classifier was first brought to attention in~\cite{Scheirer_2013_TPAMI}. It was later shown that deep learning models are also affected by the same phenomena~\cite{bendale2016towards}. Authors in~\cite{bendale2016towards} proposed a statistical solution, called OpenMax, for this problem. They converted the $c$-class classification problem into a $c+1$ problem by considering the extra class to be the open-set class. They apportioned logits of known classes to the open-set class considering spatial positioning of a query sample in an intermediate feature space. This formulation was later adopted by~\cite{ge2017generative} and \cite{neal2018open} by using a generative model to produce logits of the open-set class. Authors in \cite{yoshihashi2019classification} argued that a generative feature contain information that can benefit open-set recognition. On these grounds they considered a concatenation of a generative feature and a classifier feature when building the OpenMax layer. A generative approach was used in~\cite{oza2019c2ae} where a class conditioned decoder was used to detect open-set samples. Works of both~\cite{oza2019c2ae} and~\cite{yoshihashi2019classification} show that incorporating generative features can benefit open-set recognition. Note that open-set recognition is more challenging than the novelty detection~\cite{Poojan2020NoveltyDistribution,Poojan2020NoveltyPatch,Perera2019Noveltytransfer,perera2019learning,oza2018one} which only aims to determine whether an observed image during inference belongs to one of the known
classes.

\noindent \textbf{Self-Supervision.}  Self-supervision is an unsupervised machine learning technique where the data itself is used to provide supervision. Recent works in self-supervision introduced several techniques to improve the performance in classification and detection tasks. For example, in~\cite{selfsup}, given an anchor image patch, self-supervision was carried out by asking the network to predict the relative position of a second image patch. In~\cite{Doersch_2017_ICCV}, a multi-task prediction framework extended this formulation, forcing the network to predict a combination of relative order and pixel color. In~\cite{gidaris2018unsupervised}, the image was randomly rotated by a factor of 90 degrees and the network was forced to predict the angle of the transformed image.

\section{Background}

\noindent \textbf{Adversarial Attacks.} Consider a trained network parameterized by parameters $\theta$. Given a  data and label pair $(\textbf{x}, \textbf{y})$,  an adversarial image $\textbf{x}_{adv}$, can be produced using $\textbf{x}_{adv} = \textbf{x} + \delta$, where $\delta$ can be determined by a given white-box attack based on the models parameters. In this paper, we consider two types of adversarial attacks. 

The first attack considered is the Fast Gradient Signed Method (FGSM)~\cite{goodfellow2014explaining} where the adversarial images are formed as follows,
\begin{equation}
\textbf{x}_{adv} = {\rm Proj}_\chi (\textbf{x}+\epsilon sign(\bigtriangledown_\textbf{x} \mathcal{L}(\textbf{x}, \textbf{y}; \theta))),
\end{equation} 
where $\mathcal{L}(\cdot)$ is a classification loss. ${\rm Proj}_\chi$ denotes the projection of its element to a valid pixel value range, and $\epsilon$ denotes the size of $l_\infty$-ball. The second attack considered is Projective Gradient Descent (PGD) attacks~\cite{madry2017towards}. Adversarial images are generated in this method as follows,
\begin{equation}
\textbf{x}_{adv}^{(t+1)} = {\rm Proj}_{\zeta\cap\chi} (\textbf{x}_{adv}^{(t)}+\epsilon_{step} sign(\bigtriangledown_\textbf{x} \mathcal{L}(\textbf{x}_{adv}^{(t)}, \textbf{y}; \theta))),
\end{equation} 
where ${\rm Proj}_{\zeta\cap\chi}(\cdot)$ denotes the projection of its element to $l_\infty$-ball $\zeta$ and a valid pixel value range, and $\epsilon_{step}$ denotes a step size smaller than $\epsilon$. We use the adversarial samples of the final step $T$: $\textbf{x}_{adv} = \textbf{x}_{adv}^{(T)}$.\\

\noindent \textbf{OpenMax Classifier.} A SoftMax classifier trained for a $c$-class problem typically has $c$ probability predictions. OpenMax is an extension where the probability scores of $c+1$ classes are produced. The probability of the final class corresponds to the open-set class. Given $c$ known classes $\mathcal{K}=\{C_1, C_2, ..., C_c \}$,  OpenMax is designed to identify open-set samples by calibrating the final hidden layer of a classifier as follows:
\begin{equation}
\hat{\textbf{\textit{l}}}_i=
\begin{cases}
\textbf{\textit{l}}_i \textbf{\textit{w}}_i& (i\leq c)\\
\textbf{$\sum^{c}_{i=1}$ \textit{l}}_i (1-\textbf{\textit{w}}_i)& (i= c+1)
\end{cases},
{\rm OpenMax}_i(\textbf{x})={\rm SoftMax}_i(\hat{\textbf{\textit{l}}})
\end{equation}
where $\textbf{\textit{l}}$ denotes the logit vector obtained prior to the SoftMax operation in  a classifier, $\textbf{\textit{w}}_i$ represents the belief that $\textbf{x}_{adv}$ belongs to the known class $C_i$. Here, the class $C_{N+1}$ corresponds to the open-set class. Belief $\textbf{\textit{w}}_i$ is calculated considering the distance of a given sample to it's class mean $\mu$ in an intermediate feature space. During training, distance of all training image samples from a given class to its corresponding class mean  $\mu$ is evaluated to form a matched score distribution. Then, a Weibull distribution is fitted to the tail of the matched distribution. If the feature representation of the input in the same feature space is $\textbf{\textit{v(x)}}$, $\textbf{\textit{w}}_i$ is calculated as 
\begin{equation}
\begin{split}
\textbf{\textit{w}}_i = 1 - \max\Big(0, \frac{\sigma-{\rm rank}(i)}{\sigma}\Big)e^{\Big( -\Big(\dfrac{|\textbf{\textit{v(x)}}-\mu_i|_2}{\eta_i}\Big)^{m_i}\Big)},
\end{split}
\end{equation}
where $m_i, \eta_i$ are parameters of the Weibull distribution that corresponding to class $C_i$. $\sigma$ is hyperparameter and ${\rm rank}(i)$ is the index in the logits sorted in the descending order.

\section{Proposed Method}
\begin{figure}[t] 
	\begin{center}	
		\includegraphics[ width=1\linewidth]{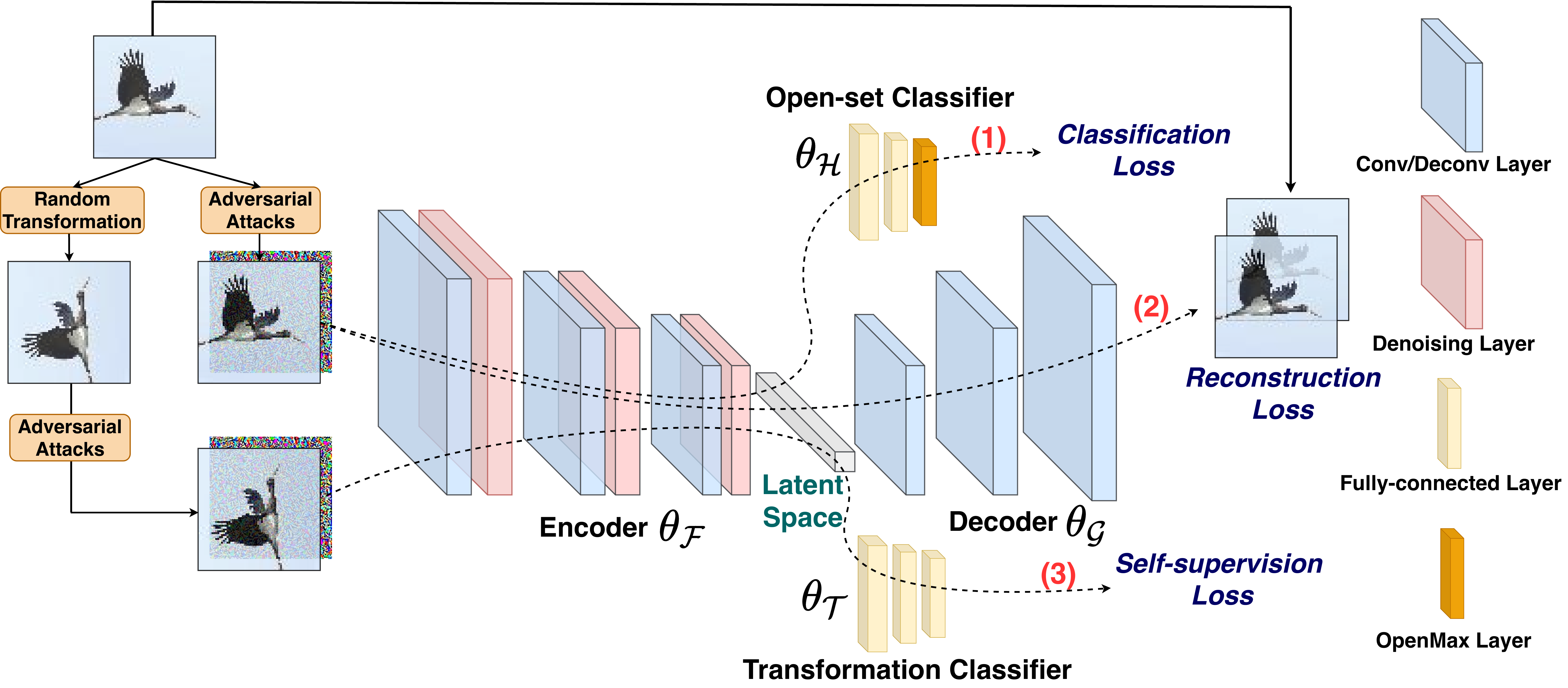}	
	\end{center}
	\caption{{Network structure of the proposed Open-Set Defense Network (OSDN). It consists of four components: encoder, decoder, open-set classifier and transformation classifier.}}
	\label{fig:nt}
\end{figure}

The proposed network consists of four CNNs: encoder, decoder, open-set classifier and transformation classifier. In Figure~\ref{fig:nt}, we illustrate the network structure of the proposed method and denote computation flow. The encoder network consists of several feature-denoising layers~\cite{xie2019feature} between the convolutional layers. Open-set classifier has no structural difference from a regular classifier. However, an OpenMax layer is added to the end of the classifier during inference. We denote this by indicating an OpenMax layer in Figure~\ref{fig:nt}.

Given an input image, first the network generates an adversarial image based on the current network parameters. This image is passed through the encoder network to obtain the latent feature. This feature is passed through the open-set classifier via path (1) to evaluate the cross entropy loss $\mathcal{L}_{cls}$. Then, the image corresponding to the obtained latent feature is generated by passing the feature through the decoder following path (2). The decoded image is used to calculate its difference to the corresponding clean image based on the reconstruction loss $\mathcal{L}_{rec}$. Finally, the input image is subjected to a geometric transform. An adversarial image corresponding to the transformed image is obtained. This image is passed through path (3) to arrive at the transformation classifier. Output of the classifier is used to calculate the cross entropy loss $\mathcal{L}_{ssd}$  considering the transform applied to the image. The network is trained end-to-end using the following loss function
\begin{equation}
\mathcal{L}_{OSDN} = \mathcal{L}_{cls} + \mathcal{L}_{rec} + \mathcal{L}_{ssd}.
\end{equation}

In the following subsections, we describe various components and computation involved in all three paths in detail.\\

\noindent \textbf{Noise-free Feature Encoding.}
The proposed network uses an encoder network to produce noise-free features. Then, the open-set classifier operating on the learned feature is used to perform classification. During training, there is no structural difference in the open-set classifier from a standard classifier. Inspired by~\cite{xie2019feature}, we embed denoising layers after the convolutional layer blocks in the encoder so that feature denoising can be explicitly carried out on adversarial samples. We adopt the Gaussian (softmax) based non-local means filter~\cite{buades2005non} as the denoising layer. Given an input feature map $m$, non-local means~\cite{buades2005non} takes a weighted mean of features in the spatial region $\mathcal{R}$ to compute a denoised feature map $g$ as follows 
\begin{equation}
g_i = \dfrac{1}{\mathcal{N}(m)}\sum\limits_{\forall j\sim\mathcal{R}}f(m_i, m_j)\cdot m_j,
\end{equation} 
where $f(m_i, m_j)$ is a feature-dependent weighting function. For the Gaussian (softmax) based version, $f(m_i, m_j)=e^{\alpha(m_i)^T\beta(m_j)/\sqrt{d}}$. $\alpha$ and $\beta$ are two $1\times1$ convolutional layers as embedding functions and $d$ corresponds to the number of channels. $\mathcal{N}(m)$ is a normalization function and $\mathcal{N}(m)=\sum_{\forall j\sim\mathcal{R}}f(m_i, m_j)$. 

Formally, we denote the encoder embedded with denoising layers as $\mathcal{F}$ parameterized by $\theta_{\mathcal{F}}$, and the classifier as $\mathcal{H}$ parameterized by $\theta_{\mathcal{H}}$. Given the labeled clean data $(\textbf{x}, \textbf{y})\sim \mathcal{I}$ from the data distribution of known classes, we can generate the corresponding adversarial images $\textbf{x}_{adv}$ on-the-fly using either FGSM or PGD attacks based on the current parameters $\theta_{\mathcal{F}}$, $\theta_{\mathcal{H}}$ using the true label $\textbf{y}$. Obtained adversarial image $\textbf{x}_{adv} $ is passed through encoder and classifier (via path (1)) to arrive at the cross-entropy loss defined as 
\begin{equation}
\mathcal{L}_{cls} = \min_{\theta_{\mathcal{F}}, \theta_{\mathcal{H}}}\mathbb{E}_{(\textbf{x}_{adv}, \textbf{y})\sim \mathcal{I}}\mathcal{L}_{CE}(\textbf{x}_{adv}, \textbf{y}; \theta_{\mathcal{F}}, \theta_{\mathcal{H}}).
\end{equation} 
By minimizing the above adversarial training loss, the trained encoder embedded with the denoising layers is able to learn a noise-free latent feature space. During inference, an OpenMax layer is added on top of the classifier. With this formulation, open-set classifier operating on the noise-free latent feature learns to predict the correct class, for both known and open-set samples, even when the input is contaminated with adversarial noise.\\

\noindent \textbf{Clean Image Generation.}
In this section, we introduce the image generation branch proposed in our method. The objective of the image generation branch is to generate noise-free images from adversarial images by taking advantage of the decoder network. This is motivated by two factors. 

First, autoencoders are widely used in the literature for image denoising applications. By forcing the autoencoder network to produce noise-free images, we are providing additional supervision to remove noise in the latent feature space. Secondly, it is a well known fact that open-set recognition becomes more effective in the presence of more descriptive features~\cite{yoshihashi2019classification}. When a classifier is trained, it models the boundary of each class. Therefore, a feature produced by a classification network only contains information that is necessary to model class boundaries. However, when the network is asked to generate noise-free images based on the latent representations, it ends up with learning generative features. As a result, features become more descriptive than in the case of a pure classifier. In fact, such generative features are used in~\cite{yoshihashi2019classification} and~\cite{oza2019c2ae} to boost the open-set recognition performance. Therefore, we argue that adding the decoder as an image generation branch can mutually benefit both open-set recognition and adversarial defense.

We pass adversarial images through path (2) as illustrated in Figure~\ref{fig:nt} to generate the decoded images. The decoder network denoted as $\mathcal{G}$ parameterized by $\theta_{\mathcal{G}}$ and the encoder network $\mathcal{F}$ are jointly optimized to  minimize the distance between the decoded images and the corresponding clean images using the following loss
\begin{equation}
\mathcal{L}_{rec} = \min_{\theta_{\mathcal{F}}, \theta_{\mathcal{G}}}\mathbb{E}_{(\textbf{x}, \textbf{x}_{adv})\sim \mathcal{I}} \|\textbf{x}-{\mathcal{G}}({\mathcal{F}}(\textbf{x}_{adv})) \|_2^2.
\end{equation}

\noindent \textbf{Self-supervised Denoising.}
Finally, we propose to use self-supervision as a means to further increase the informativeness and robustness of the latent feature space. Self-supervision is a machine learning technique that is used to learn representations in the absence of labeled data. In our work  we adopt rotation-based self-supervision proposed in~\cite{gidaris2018unsupervised}. In~\cite{gidaris2018unsupervised}, first, a random rotation from a finite set of possible rotations is applied to an image. Then, a classifier is trained on top of a latent feature vector to automatically recognize the applied rotation. 

In our approach, similar to \cite{gidaris2018unsupervised}, we first generate a random number $r \in \{0, 1, 2, 3\}$ as the rotation ground-truth and transform the input clean image $\textbf{x}$ by rotating it with $ 90^\circ r$ degrees. Then, based on the rotated clean image, we generate a rotated adversarial image $\textbf{x}^{\mathcal{T}}_{adv}$ on-the-fly using either FGSM or PGD attack based on the current network parameters and rotation ground-truth $r$, which is passed through the transformation classifier to generate the cross-entropy loss with respect to the ground-truth $r$. We denote the transformation classifier as $\mathcal{T}$ parameterized by $\theta_{\mathcal{T}}$ and formulate the adversarial training loss function for self-supervised denoising as follows
\begin{equation}
\mathcal{L}_{ssd} = \min_{\theta_{\mathcal{F}}, \theta_{\mathcal{T}}}\mathbb{E}_{\textbf{x}^{\mathcal{T}}_{adv}\sim \mathcal{I}} \mathcal{L}_{CE}(\textbf{x}^{\mathcal{T}}_{adv}, r; \theta_{\mathcal{F}}, \theta_{\mathcal{T}}).
\end{equation} 

There are multiple reasons why we use self-supervision in our method. When a classifier learns to differentiate between different rotations, it learns to pay attention to object structures and orientations of known classes. As a result, when self-supervision is carried out in addition to classification, the underlying feature space learns to represent additional information that was not considered in the case of a pure classifier. Therefore, self-supervision enhances the informativeness of the latent feature space which would directly benefit the open-set recognition process. On the other hand, since we use adversarial images for self-supervision, this operation directly contributes towards learning the denoising operator in the encoder. It should also be noted that recent work~\cite{hendrycks2019using} has found that self-supervised learning contributes towards robustness against adversarial samples. Therefore, we believe that addition of self-supervision benefits both open-set detection and adversarial defense processes.\\

\noindent \textbf{Implementation Details.} 
We adopt the structure of Resnet-18~\cite{Kaiming_Resnet_CVPR2016}, which has four main blocks, for the encoder network.  Denoising layers are embedded after each main blocks in the encoder. For the decoder, we use the decoder network proposed in~\cite{neal2018open} with three transpose-convolution layers for conducting experiments with the SVHN and CIFAR10 dataset. Four transpose-convolution layers are used for conducting experiments with the TinyImageNet dataset. For both open-set classifier and transformation classifier, we use a single fully connected layers. We use the Adam optimizer~\cite{kingma2014adam}  for the optimization with a learning rate of 1e-3. We carried out model selection considering the trained model that has produced  the best closed-set accuracy  on the validation set.  We use the iteration = 5 for the PGD attacks and $\epsilon=0.3$ for the FGSM attacks in both adversarial training and testing.

\section{Experimental Results}
In order to assess the effectiveness of the proposed method, we carry out experiments on three multiple-class classification datasets. In this section, we first describe datasets, baseline methods and the protocol used in our experiments. We evaluate our method and baselines in the task of open-set recognition under adversarial attacks. To further validate the effectiveness of our method, additional experiments in the task of out-of-distribution detection under adversarial attacks are conducted.  We conclude the section by presenting an ablation study and various visualizations with a brief analysis of the results.

\subsection{Datasets}

The evaluation of our method and other state-of-the-arts are conducted on three standard images classification datasets for open-set recognition:

\noindent \textbf{SVHN and CIFAR10.} Both CIFAR10~\cite{cifar10Hinton} and SVHN~\cite{netzer2011reading} are classification datasets with 10 classes with images of size 32$\times$32.  Street-View House Number dataset (SVHN) contains house number signs extracted from Google Street View. CIFAR10 contains images from four vehicle classes and six animal classes. We randomly split 10 classes into 6 known classes and 4 open-set classes to simulate open-set recognition scenario. We consider three randomly selected splits for testing\footnote{Details about known classes present in each split can be found in supplementary materials.}. 

\noindent \textbf{TinyImageNet.} TinyImageNet contains a sub-set of 200 classes selected from the ImageNet dataset~\cite{deng2009imagenet} with image size  of 64$\times$64. 20 classes are randomly selected to be known and the remaining 180 classes are chosen to be open-set classes. We consider three randomly chosen splits for evaluation.

\subsection{Baseline Methods.}
We consider the following two recently proposed adversarial defense methods as baselines: \textbf{Adversarial Training}~\cite{madry2017towards} and \textbf{Feature Denoising}~\cite{xie2019feature}. We add an OpenMax layer in the last hidden layer during testing for both baselines to facilitate a fair comparison in open-set recognition. Moreover, to evaluate the performance of a classifier without a defense mechanism, we train a Resnet-18 network on clean images obtained from known classes and add an OpenMax layer during testing. We test this network using clean images for inference and we denote this test case by \textbf{clean}. Furthermore, we  test this model with adversarial images, which is denoted as \textbf{adv on clean}.

\subsection{Quantitative Results}

\begin{table}[!htb]
	\scriptsize
	\renewcommand{\arraystretch}{1}
	\centering
	\caption{ Adversarial Defense: Closed-set accuracy. }
	\begin{tabular}{c|c|c|c|c|c|c}
		\hline
		\multirow{2}{*}{Method}    & \multicolumn{2}{c|}{\textbf{SVHN}} & \multicolumn{2}{c|}{\textbf{CIFAR-10}} & \multicolumn{2}{c}{\textbf{TinyImageNet}} \\ \cline{2-7} 
		& FGSM         & PGD        & FGSM           & PGD          & FGSM             & PGD            \\ \hline
		clean                & 96.0$\pm$0.6 &96.0$\pm$0.6  &93.1$\pm$1.8 &93.1$\pm$1.8   &56.8$\pm$3.6 & 56.8$\pm$3.6                \\ 
		adv on clean         & 41.6$\pm$3.2 &39.3$\pm$1.8  &31.8$\pm$4.5 &13.0$\pm$4.0   &11.2$\pm$2.6 & 4.4$\pm$0.8                \\ \hline
		adversarial training & 88.5$\pm$2.7 &75.8$\pm$2.5  &87.3$\pm$1.1 &72.4$\pm$4.6   &66.6$\pm$1.2 & 40.3$\pm$2.3               \\ 
		feature denoising    & 86.9$\pm$3.7 &75.5$\pm$2.6  &87.4$\pm$2.3 &72.5$\pm$4.5   &64.5$\pm$1.3 & 39.3$\pm$3.0               \\ \hline
		\textbf{ours}                 & \textbf{89.3}$\pm$0.7 &\textbf{77.9}$\pm$1.6  &\textbf{88.2}$\pm$2.9 &\textbf{74.2}$\pm$4.3   &\textbf{75.1}$\pm$7.9 & \textbf{41.6}$\pm$2.2               \\ \hline
	\end{tabular}
	\label{tbl:osacc}
\end{table}

\noindent \textbf{Open-set Recognition.}  In conventional open-set recognition, the model is required to perform two tasks. First, it should be able to detect open-set samples effectively. Secondly, it should be able to perform correct classification on closed set samples. In order to evaluate the open-set defense performance, we take these two factors into account. In particular, following previous open-set works~\cite{neal2018open}, we use area under the ROC curve (AUC-ROC) to evaluate the performance on identifying open-set samples under adversarial attacks. In order to evaluate the closed-set accuracy, we calculate prediction accuracy by only considering known-set samples in the test set. In our experiments, both known and open-set samples were subjected to adversarial attacks prior to testing. During our experiments we consider FGSM and PGD attacks to attack the model. We generated adversarial samples from known classes using the ground-truth labels, while we generated the adversarial samples from open-set classes based on model's prediction.  

We tabulate the obtained performance for closed-set accuracy and open-set detection in Tables~\ref{tbl:osacc} and~\ref{tbl:osauc}, respectively.  We note that, networks trained on clean images produce very high recognition performance  for clean images under both scenarios. However, when the adversarial noise is present, both open-set detection and closed-set classification performance drops significantly. This validates that current adversarial attacks can easily fool an open-set recognition method such as OpenMax, and thus OSAD is a critical research problem. Both baseline defense mechanisms considered are able to improve the recognition on both known and open-set samples.  It can be observed from Tables~\ref{tbl:osacc} and~\ref{tbl:osauc}, that the proposed method obtains the best open-set detection performance and closed-set accuracy compared to all considered baselines across all three datasets. In particular, the proposed method has achieved about $7\%$ improvement in open-set detection on the SVHN dataset compared to the other baselines. On other datasets, this improvement varies between $1-5\%$. The proposed method is also able to perform better in terms of closed-set accuracy compared to the baselines consistently across datasets.

It is interesting to note that methods involving adversarial training perform better than the baseline of clean image classification under FGSM attacks on the TinyImageNet dataset. This is because only 20 classes from the TinyImageNet dataset are selected for training and each class has only 500 images.  When a small dataset is used to train a model with large number of parameters, it is easier for the network to overfit to the training set. Such network observes variety of data in the presence of adversarial training.  Therefore model reaches a more generalizable optimization solution during training. 

\begin{table}[!htbp]
	\scriptsize
	\renewcommand{\arraystretch}{1}
	\centering
	\caption{ Open Set Classification: Area under the ROC curve. }
	\begin{tabular}{c|c|c|c|c|c|c}
		\hline
		\multirow{2}{*}{Method}    & \multicolumn{2}{c|}{\textbf{SVHN}} & \multicolumn{2}{c|}{\textbf{CIFAR-10}} & \multicolumn{2}{c}{\textbf{TinyImageNet}} \\ \cline{2-7} 
		& FGSM         & PGD        & FGSM           & PGD          & FGSM             & PGD            \\ \hline
		clean                &  91.3$\pm$2.4 & 91.3$\pm$2.4 &  81.2$\pm$2.9   & 81.2$\pm$2.9   & 59.5$\pm$0.8          &59.5$\pm$0.8                \\ 
		adv on clean       &  56.4$\pm$1.2 & 54.1$\pm$2.9 &  51.5$\pm$2.8   & 45.5$\pm$0.5   & 47.9$\pm$2.7          &48.6$\pm$1.3                \\ \hline
		adversarial training &  61.4$\pm$8.0 & 65.2$\pm$4.0 &  75.2$\pm$1.2   & 68.7$\pm$3.2   & 65.1$\pm$8.1          &56.5$\pm$0.9                \\ 
		feature denoising    &  64.5$\pm$14.7  &64.9$\pm$4.2&  76.9$\pm$3.7   & 69.8$\pm$2.4   & 65.3$\pm$5.1          &56.1$\pm$1.6               \\ \hline
		\textbf{ours}                 &  \textbf{71.4}$\pm$4.2  & \textbf{71.6}$\pm$2.6&  \textbf{79.1}$\pm$1.0   & \textbf{70.6}$\pm$1.7   & \textbf{70.8}$\pm$5.1          &\textbf{58.2}$\pm$1.9                \\ \hline
	\end{tabular}
	\label{tbl:osauc}
\end{table}

\noindent \textbf{Out-of-distribution detection.}
In this sub-section, we evaluate the performance of the proposed method on the out-of-distribution detection (OOD)~\cite{hendrycks2016baseline} problem on CIFAR10 using the protocol  described in~\cite{yoshihashi2019classification}. We considered all classes in CIFAR10 as known-classes and consider test images from ImageNet and LSUN dataset~\cite{yu2015lsun} (both cropped and resized) as out-of-distribution images~\cite{liang2017enhancing}. We tested the performance of adversarial images by creating adversarial images using the PGD attacks for both known and OOD data. We generated adversarial samples from the known classes using the ground-truth labels, while we generated adversarial samples from the OOD class  based on model's prediction. We evaluated the performance of the model on adversarial samples based on macro-averaged F1 score. We used OpenMax layer with threshold $0.95$ when assigning open-set labels to the query images, In Table~\ref{tbl:f1}, we tabulate the OOD detection performance across all four cases considered for both baselines as well as the proposed method. As evident from Table~\ref{tbl:f1}, the proposed method outperforms baseline methods in all test cases in the ODD task. This experiment further verifies the effectiveness of our method to identify samples from open-set classes even under the adversarial attacks.

\begin{table}[!htb]
	\scriptsize
	\renewcommand{\arraystretch}{1}
	\centering
	\caption{Performance of out-of-distribution object detection on the CIFAR10 dataset.}	
	\begin{tabular}{c|c|c|c|c}  
		\hline
		Detector             & ImageNet-Crop  & ImageNet-Resize & LSUN-Crop      & LSUN-Resize    \\ \hline
		clean                & 78.9          & 76.2           & 82.1          & 78.7          \\ 
		adv on clean         & 4.7          & 4.4           & 7.3          & 3.8          \\ \hline
		adversarial training & 35.2          & 34.5           & 35.0           & 34.7          \\ 
		feature denoising    & 43.2          & 41.0            & 43.5          & 41.2          \\ \hline
		\textbf{ours}        & \textbf{46.5} & \textbf{44.8}  & \textbf{47.1} & \textbf{44.2} \\ \hline
	\end{tabular}
	\label{tbl:f1}
\end{table}

\subsection{Ablation Study}
The proposed network consists of four CNN components. In this sub-section we investigate the impact of each network component to the overall performance of the system. To validate the effectiveness of various parts integrated in our proposed network, this section conducts the ablation study in our network using the CIFAR10 dataset for the task of open-set recognition. Considered cases and the corresponding results obtained for each case are tabulated Table~\ref{tblab} (C-accuracy means closed-set accuracy). From Table~\ref{tblab}, it can be seen that compared to normal adversarial training with an encoder, embedding the denoising layers helps to improve the open-set classification performance. Moreover, as evident from  Table~\ref{tblab}, adding a denoising layer to perform feature denoising and adding self-supervision both have resulted in improved performance.  The proposed method that integrates all these components has the best performance, which shows that added components complement each other to produce better performance for both adversarial defense and open-set recognition.

\begin{table}[!htb]
	\scriptsize
	\renewcommand{\arraystretch}{1}
	\centering
	\caption{Results corresponding to the ablation study. }
	\begin{tabular}{c|p{2cm}<{\centering}|p{2cm}<{\centering}}
		\hline
		\multirow{2}{*}{Method} & \multicolumn{2}{c}{\textbf{CIFAR-10}} \\ \cline{2-3} 
		& AUC-ROC       & C-accuracy       \\ \hline
		clean                   			&83.7        &  92.7          \\ 
		adv on clean            			&45.9        &  8.6          \\ \hline
		Encoder                      		 &   66.1                        &  69.9          \\ 
		Encoder+Denoising Layer                    &    68.5              &   70.4         \\ 
		Encoder+Decoder                     &67.3                  &68.8          \\ 
		Encoder+Decoder+Denoising Layer                  &  68.2                         &   70.6         \\ 
		Encoder+Decoder+self-supervised Denoising                &   68.5     &  71.9          \\ \hline
		\textbf{ours}                    &    \textbf{69.6}        &   \textbf{72.8}         \\ \hline
	\end{tabular}
	\label{tblab}
\end{table}

\begin{figure*}[htb]
	\centering
	\begin{minipage}[t]{0.47\linewidth}
		\centering
		\includegraphics[width=1\linewidth]{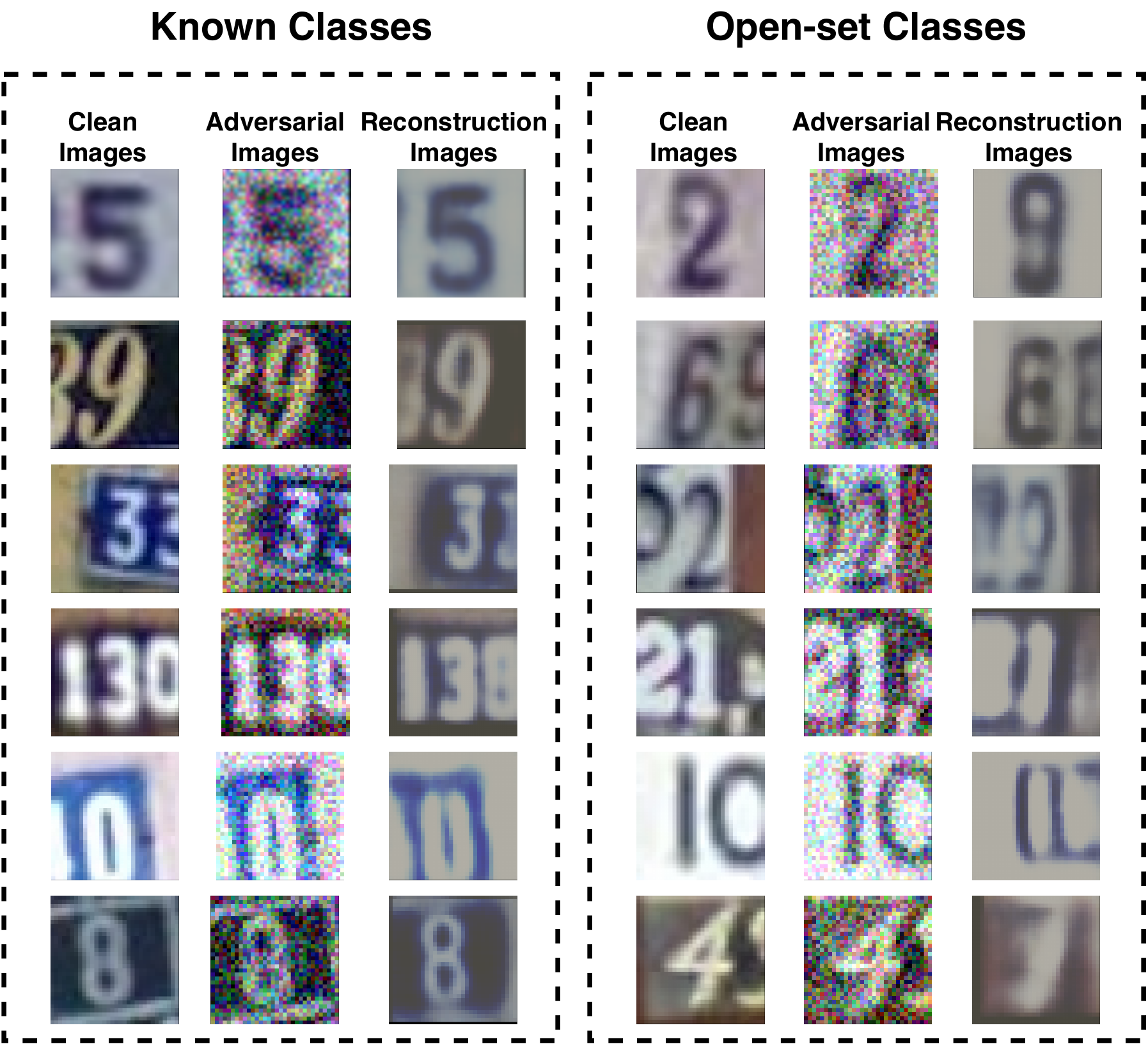}
		\caption{Visualization of input clean images, corresponding adversarial images, and the reconstructed images from the decoder.}
	\label{fig:vis_rec}
	\end{minipage}
	\begin{minipage}[t]{0.35\linewidth}
		\centering
		\includegraphics[width=1\linewidth]{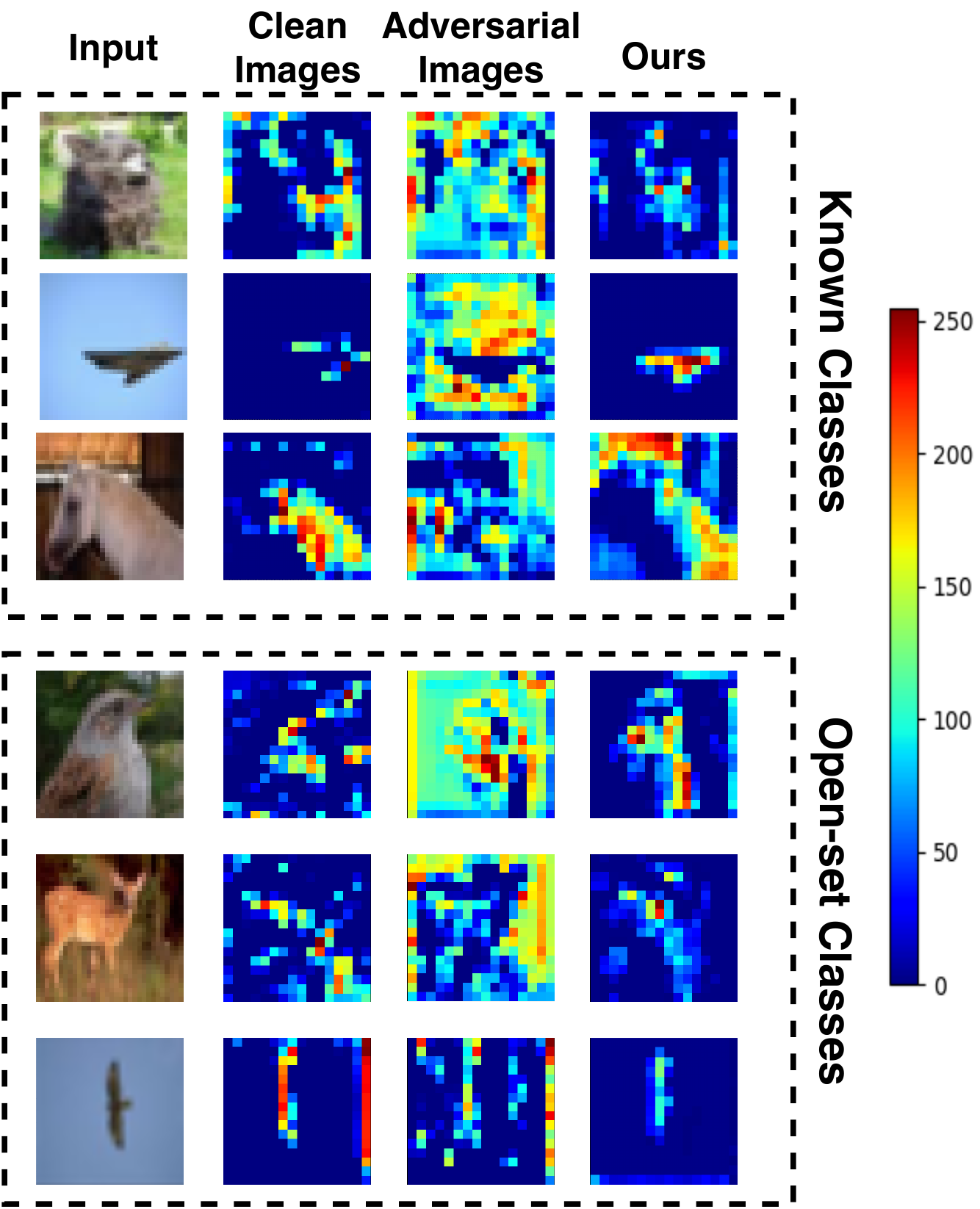}  
		\caption{Feature map visualization in the res$_2$ block of Resnet-18 and the encoder of proposed network.}
	\label{fig:vis_featmap}
	\end{minipage}
\end{figure*}

\subsection{Qualitative Results}
In this section, we visualize the results of the denoising operation and obtained features in a 2D plane to qualitatively analyze the performance of the proposed method. For this purpose, we first consider the  SVHN dataset. Figure~\ref{fig:vis_rec} shows a set of clean images, corresponding PGD attacks adversarial images and images obtained when the latent feature is decoded under the proposed method. We have indicated known and open-set sample visualizations in two columns.  From the image samples shown in Figure~\ref{fig:vis_rec}, it can be observed that image noise has been removed in both open-set and known-class images. However, the reconstruction quality is superior for the known-class samples compared to the open-set images. Reconstructions of open-set samples look blurry and structurally different. For example, the image of digit 2 shown in the first row, looks similar to the digit 9 once reconstructed. 

\begin{figure}[!htb] 
	
	\begin{center}
		
		\includegraphics[height=3.5cm, width=0.9\linewidth]{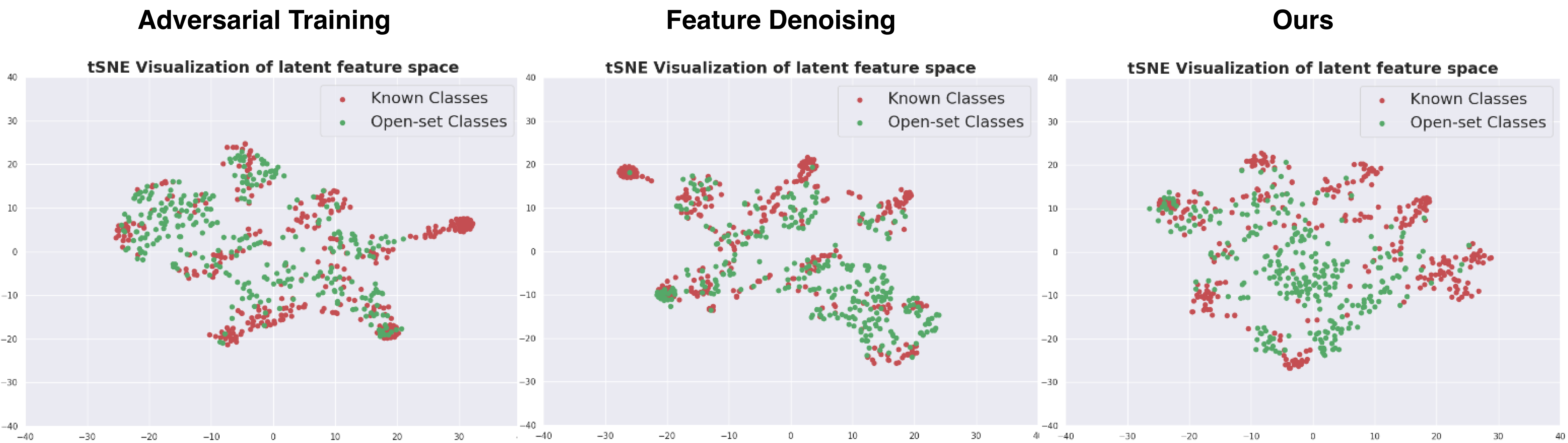}
		
	\end{center}
	\caption{{tSNE visualization of the latent feature space corresponding to our method and two baselines.}}
	\label{fig:tsne}
\end{figure}

In Figure~\ref{fig:tsne} we visualize latent features obtained in the proposed method along with two other baselines using tSNE visualization~\cite{maaten2008visualizing} . As shown in Figure~\ref{fig:tsne}, most of open-set features lie away from the known-set feature distribution based on our method. This is why the proposed method is able to obtain good open-set detection performance. On the other hand, it can be observed from Figure~\ref{fig:tsne} that there is more overlap between the two types of features in all baseline methods. When open-set features lie away from the data manifold of known set classes, the reconstruction obtained through the decoder network tends to be poor. Therefore, the tSNE plot justifies why the reconstruction of our method was poor for open-set samples in Figure~\ref{fig:vis_rec}. As such, Figure~\ref{fig:vis_rec} and Figure~\ref{fig:tsne} mutually verify the effectiveness of our method for defending adversarial samples and identifying open-set samples simultaneously.

Moreover, we visualize randomly selected feature maps of the second residual block from the trained Resnet-18~\cite{Kaiming_Resnet_CVPR2016} and the encoder of the proposed OSDN network applied on clean images and on its adversarially perturbed counterpart in the CIFAR10 dataset. From Figure~\ref{fig:vis_featmap}, it can be observed that compared to Resnet-18, the proposed network is able to reduce adversarial noise significantly in feature maps of adversarial images in both known and open-set classes. This further demonstrates that the proposed network indeed carries out the feature denoising through the embedded feature denoising layers.

\section{Conclusion}
In this paper, we studied a novel research problem -- Open-set Adversarial Defense (OSAD). We first showed that existing adversarial defense mechanisms do not generalize well to open-set samples. Furthermore, we showed that even open-set classifiers can be easily attacked using the existing attack mechanisms. We proposed an Open-Set Defense Network (OSDN) with the objective of producing a model that can detect open-set samples while being robust against adversarial attacks. The proposed network consisted of feature denoising operation, self-supervision operation and a denoised image generation function. We demonstrated the effectiveness of the proposed method under both open-set and adversarial attack settings on three publicly available classification datasets. Finally, we showed that proposed method can be deployed for out-of-distribution detection task as well. 

\section*{Acknowledgments}
This work is partially supported by Research Grants Council (RGC/HKBU12200518), Hong Kong. Vishal M. Patel was supported by the DARPA GARD Program HR001119S0026-GARD-FP-052.
%
%
\bibliographystyle{splncs04}
\bibliography{egbib}
\end{document}